\documentclass[conference]{IEEEtran}
\IEEEoverridecommandlockouts
\usepackage{cite}
\usepackage{amsmath,amssymb,amsfonts}
\usepackage{algorithmic}
\usepackage{graphicx}
\usepackage{textcomp}
\usepackage{xcolor}
\def\BibTeX{{\rm B\kern-.05em{\sc i\kern-.025em b}\kern-.08em
    T\kern-.1667em\lower.7ex\hbox{E}\kern-.125emX}}
\begin{document}

\title{Resilient Routing: Risk-Aware Dynamic Routing in Smart Logistics via Spatiotemporal Graph Learning}

\author{
    \IEEEauthorblockN{
        Zhiming Xue\IEEEauthorrefmark{2}\textsuperscript{*},
        Sichen Zhao\IEEEauthorrefmark{2},
        Yalun Qi\IEEEauthorrefmark{3},
        Xianling Zeng\IEEEauthorrefmark{2} and
        Zihan Yu\IEEEauthorrefmark{4}
    }

    \IEEEauthorblockA{\IEEEauthorrefmark{2}College of Engineering, Northeastern University, Boston, USA}
    \IEEEauthorblockA{\IEEEauthorrefmark{3}Khoury College of Computer Sciences, Northeastern University, Boston, USA}
    \IEEEauthorblockA{\IEEEauthorrefmark{4}College of Professional Studies, Northeastern University, Boston, USA}

    \IEEEauthorblockA{
        Emails: \{xue.zh\textsuperscript{*}, zhao.siche, qi.yal, zeng.xian, yu.zihan1\}@northeastern.edu
    }
}

\maketitle

\begin{abstract}
With the rapid development of the e-commerce industry, the logistics network is experiencing unprecedented pressure. Traditional static routing strategies are often unable to tolerate traffic congestion and fluctuating retail demand. In this paper, we propose a Risk-Aware Dynamic Routing(RADR) framework which integrates  Spatiotemporal Graph Neural Networks (ST-GNN) with combinatorial optimization. We first construct a logistics topology graph by combining spatial clustering with trajectory-induced connectivity learning to capture implicit road network constraints. Subsequently, a hybrid deep learning model combining Graph Convolutional Network (GCN) and Gated Recurrent Unit (GRU) is adopted to extract spatial correlations and temporal dependencies for predicting future congestion risks. These prediction results are then integrated into a dynamic edge weight mechanism to perform path planning. We evaluated the framework on the Smart Logistics Dataset 2024, which contains real-world Internet of Things(IoT) sensor data. The experimental results show that the RADR algorithm significantly enhances the resilience of the supply chain. Particularly in the case study of high congestion scenarios, our method reduces the potential congestion risk exposure by 17.6\% while only increasing the transportation distance by 2.1\%.

\end{abstract}

\begin{IEEEkeywords}
Smart Logistics, Graph Neural Network, Dynamic Routing, Spatiotemporal modeling, Supply Chain Resilience
\end{IEEEkeywords}

\section{Introduction}
Logistics and supply chain management serve as the backbone of the modern global economy. However, the "Last-Mile Delivery" sector faces significant challenges due to the stochastic nature of urban traffic and the volatility of retail demand. Traditional path planning algorithms, such as Dijkstra and A* algorithms, typically rely on static physical distances, assuming that the network topology and edge weights remain constant over time. This assumption often leads to suboptimal routing decisions, guiding vehicles into congested areas, which results in delivery delays and increased operational costs.

To address these limitations, recent research has begun to explore data-driven approaches. A comprehensive survey by Bogyrbayeva et al. \cite{b1} highlights that machine learning-based solvers effectively address the scalability issues of traditional heuristics in dynamic environments. While deep learning models like Long Short-Term Memory (LSTM) networks have shown success in time-series traffic prediction, they often treat traffic data as isolated sequences, neglecting the complex spatial dependencies within the logistics network. In contrast, Graph Neural Networks (GNNs) excel at modeling through topological structures but often lack end-to-end integration with downstream decision-making processes. Recent studies have demonstrated the superiority of Graph Neural Networks (GNNs) in modeling non-Euclidean spatial dependencies in intelligent transportation systems. \cite{b2} Representative spatiotemporal graph forecasting models such as DCRNN \cite{b3} and STGCN \cite{b4} have demonstrated strong performance by jointly modeling graph-structured spatial dependencies and temporal dynamics. Specifically, in supply chain scenarios, GNNs are increasingly used to bridge the gap between demand forecasting and logistics planning. From a broader industrial perspective, recent studies have highlighted the transformative role of generative AI in accelerating industrial digitalization and operational intelligence across sectors, emphasizing its growing impact on complex decision-making and optimization tasks \cite{b5}. 

In this paper, we move beyond treating prediction and optimization as isolated tasks. We present a unified framework that closes the loop, using real-time spatiotemporal forecasts to directly drive dynamic routing decisions. Our work focuses on four key architectural innovations:

\begin{enumerate}
\item We introduce a data-driven discretization method using K-Means, effectively transforming unstructured, continuous GPS trajectories into a structured logistics graph that GNNs can digest.
\item We design a hybrid ST-GNN architecture that fuses GCN and GRU layers, allowing the system to simultaneously capture the spatial topology of the road network and the temporal evolution of traffic waves.
\item We implement a risk-aware optimization mechanism that translates raw congestion scores into dynamic routing penalties, enforcing risk-averse behavior in path planning.
\item We validate the system on the Smart Logistics Dataset 2024, demonstrating that our approach successfully navigates the trade-off between minimizing distance and maximizing supply chain resilience.
\end{enumerate}

\section{Related Work}

\subsection{Traffic and Demand Prediction}

Traffic prediction has undergone a development process from classical statistical methods to advanced deep learning technologies. Research in early stage, mainly relied on ARIMA (Autoregressive Integral Moving Average Model) and its variants. These methods are effective for linear time series data, but often fall short when dealing with complex nonlinear dependencies. With the rise of deep learning, Recurrent Neural Networks (RNNs), especially LSTM (Long Short-Term Memory) networks , have become standard methods due to their ability to effectively capture the temporal dynamics of traffic flow. In recent years, spatiotemporal graph models such as STGCN has been proposed to jointly capture road-network topology and temporal evolution, achieving strong performance in traffic forecasting tasks. Recent studies have also demonstrated that transformer-based architectures can effectively capture long-term dependencies in multivariate time-series forecasting, providing robust predictive performance under complex temporal dynamics \cite{b6}. Notably, Kong et al. \cite{b7} recently proposed pivotal graph learning strategies at AAAI 2024, further confirming the dominance of GNNs in modeling complex spatiotemporal dependencies.

\subsection{Logistics Routing Optimization}

The vehicle routing problem (VRP) has long served as a fundamental formulation for logistics and transportation optimization, aiming to determine efficient delivery routes under various operational constraints such as travel cost, time windows, and vehicle capacity. Due to its combinatorial nature and NP-hard complexity, VRP has motivated extensive research across operations research and intelligent transportation systems.

Traditional approaches to VRP typically rely on heuristic and metaheuristic optimization techniques, including ant colony optimization (ACO), genetic algorithms (GA), and their numerous variants. These methods have demonstrated strong performance in static or moderately sized problem instances; however, they often require extensive parameter tuning and exhibit limited adaptability when facing dynamic demand, real-time traffic variation, or large-scale network changes.

More recently, learning-based routing methods have emerged as a promising alternative. Almasan et al. \cite{b8} explored the integration of graph neural networks with reinforcement learning for routing optimization, demonstrating their capability to model complex network structures and generalize across varying graph topologies. By representing logistics systems as graphs, GNN-based approaches enable data-driven routing strategies that better adapt to evolving operational conditions, providing new opportunities for dynamic and large-scale logistics optimization. Similar graph-based learning paradigms have also been applied to complex financial transaction workflows, where relational graph convolutional networks were employed to model heterogeneous entities and interactions, demonstrating the effectiveness of GNNs in real-world decision-support systems \cite{b9}.

\subsection{Research Gap}

Despite considerable progress in both areas mentioned above, most of the existing literature treats traffic prediction and route optimization as two separate tasks, predicting traffic first and then feeding the result back to the solver as an afterthought. In contrast, our work proposes an end-to-end framework that directly integrates the risk prediction of ST-GNN into the dynamic decision-making process, thus enabling proactive instead of reactive routing strategies.

\section{METHODOLOGY}

This section presents the proposed Risk-Aware Dynamic Routing (RADR) framework in a formal and implementation-oriented manner. The objective is to transform raw, noisy IoT logistics data into reliable routing decisions by tightly coupling spatiotemporal risk prediction with classical graph-based optimization.

\subsection{Graph Topology Construction}
Real-world logistics trajectories are continuous and noisy, which do not naturally conform to the discrete graph structures required by graph neural networks. We first apply K-Means clustering on all GPS points to discretize the service area into $N=10$ functional zones. Each cluster centroid is treated as a node $v \in V$.

The value of $N$ is selected to balance spatial granularity and computational efficiency. Preliminary experiments indicate that smaller values lead to over-smoothing of traffic patterns, while larger values increase graph sparsity and model instability. The resulting graph provides a compact yet expressive structural backbone for subsequent spatiotemporal modeling.

To construct a realistic base graph, we derive connectivity directly from historical GPS trajectories rather than relying solely on geometric proximity. Specifically, trajectory records are grouped by the real \textit{Truck\_ID}, and only temporally consecutive GPS records within each truck trajectory are used to derive inter-zone transitions. Let $C_{ij}$ denote the number of observed transitions from zone $i$ to zone $j$ across all trajectories. A directed adjacency matrix $A \in \mathbb{R}^{N \times N}$ is then obtained by row-normalizing the transition counts:
\begin{equation}
\mathbf{A}_{ij} = \frac{C_{ij}}{\sum_{k} C_{ik} + \epsilon}
\label{eq:row_norm_adj}
\end{equation}

To suppress spurious connections caused by noise or rare movements, edges with $A_{ij} < \tau$ are pruned, where $\tau=0.01$ in our experiments. The resulting trajectory-induced directed graph captures asymmetric mobility patterns between zones and provides a more realistic abstraction of the underlying logistical network. The normalized transition weights $A_{ij}$ are further used as edge weights in the GCN layer.

\subsection{Spatiotemporal Risk Prediction Model}
\begin{figure}[htbp]
  \centering
  \includegraphics[width=0.85\columnwidth]{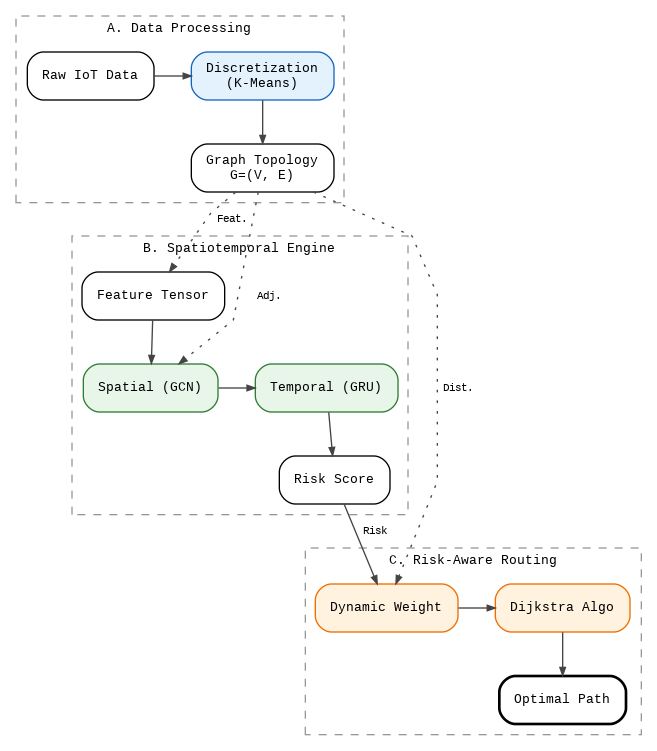}
  \caption{The proposed RADR framework architecture}
  \label{fig:system_architecture}
\end{figure}

With the static topology set, we needed a way to understand the dynamics flowing through it. Standard time-series models like LSTM are great at seeing what happened yesterday, but they are blind to what's happening nearby. A traffic jam in one hub almost always bleeds into its neighbors, and we needed a model that could capture this "spatial leakage."

We solved this by designing a hybrid Spatiotemporal Graph Neural Network (ST-GNN). To capture the complex spatiotemporal dependencies of the road network, we adopt a Spatiotemporal Graph Neural Network (ST-GNN) architecture, which has been proven effective in capturing traffic volatility in urban environments\cite{b3,b4}. The workflow operates in two stages:

\begin{enumerate}
    \item Spatial Modeling via Graph Convolution: At time step $t$, node-level traffic features are first processed by a Graph Convolutional Network (GCN) layer to aggregate spatial information from neighboring nodes. The layer-wise propagation rule is defined as:
\begin{equation}
\mathbf{H}^{(l+1)} = \sigma\left(\tilde{\mathbf{D}}^{-\frac{1}{2}}\tilde{\mathbf{A}}\tilde{\mathbf{D}}^{-\frac{1}{2}}\mathbf{H}^{(l)}\mathbf{W}^{(l)}\right)
\label{eq:gcn}
\end{equation}
where $\tilde{A} = A + I$ denotes the adjacency matrix with self-loops, $\tilde{D}$ is the corresponding degree matrix, $W^{(l)}$ is the learnable weight matrix, and $\sigma(\cdot)$ is a nonlinear activation function. This operation enables each node to incorporate localized traffic conditions from its immediate neighborhood, thereby capturing spatial congestion diffusion.
    \item Temporal Modeling via GRU: The spatially enriched node representations are then passed to a Gated Recurrent Unit (GRU) to model temporal dependencies. Given a historical observation window of length $T=10$, the GRU processes the sequence of graph embeddings and outputs a one-step-ahead congestion risk prediction $\hat{y}_{t+1}$. We adopt GRU instead of LSTM due to its reduced parameter count and faster convergence, which is well-suited for the dataset scale considered in this study. The GRU effectively captures evolving traffic dynamics such as congestion buildup and dissipation across time.
\end{enumerate}

\subsection{Risk-Aware Dynamic Routing}

The predicted congestion risk is directly integrated into the routing decision process through a dynamic edge weighting mechanism. Rather than modifying the routing algorithm itself, we embed risk awareness into the cost function of a standard Dijkstra shortest-path solver.

For each edge $(u, v)$, the dynamic weight is defined as:
\begin{equation}
W_{dyn}(u, v) = \text{dist}(u, v) \times (1 + \lambda \cdot \text{Risk}_{avg})\label{eq:dynamic_weight}
\end{equation}
where $dist(u, v)$ is the physical distance between nodes, $Risk_{avg}$ is the predicted average congestion risk associated with the edge, and $\lambda$ is a penalty coefficient controlling risk sensitivity. By inflating the effective cost of high-risk edges, the routing algorithm is naturally guided toward safer alternatives.

The optimal route $P^*$ from source node $u$ to destination node $v$ is obtained by minimizing the cumulative dynamic cost:
\begin{equation}
P^* = \arg\min_{P \in \mathcal{P}_{uv}} \sum_{(i,j) \in P} W_{dyn}(i, j)
\label{eq:objective}
\end{equation}
where $\mathcal{P}_{uv}$ denotes the set of all feasible paths between $u$ and $v$. This formulation allows the system to explicitly trade marginal increases in travel distance for significant reductions in congestion risk, thereby enhancing route reliability under uncertain traffic conditions.

To quantitatively evaluate route-level risk exposure, we define a path-level Risk Score by aggregating the predicted edge-level congestion risks along the selected route. Specifically, the Risk Score of a path P is computed as the cumulative sum of the predicted average risk values over all edges on the path:

\begin{equation}
Risk(P) = \sum_{(i,j) \in P} Risk_{avg}(i,j)
\label{eq:objective}
\end{equation}

This metric reflects the total congestion risk exposure encountered along a delivery route and is used for comparative analysis between routing strategies in Table II. Lower Risk Scores indicate safer routes with reduced likelihood of congestion-induced delays.

Overall, the proposed RADR framework establishes an end-to-end pipeline that tightly couples spatiotemporal risk prediction with classical graph optimization, enabling proactive and resilient routing decisions in smart logistics networks.

\section{Experiments}

\subsection{Dataset \& Settings}

All experiments are conducted using the Smart Logistics Supply Chain Dataset (2024), a publicly available dataset hosted on Kaggle, which contains real-world logistics and supply chain operational records collected throughout 2024.

The dataset provides fine-grained, timestamped records related to logistics activities, including GPS trajectories, shipment identifiers, temporal delivery events, traffic-related indicators, and delay labels, reflecting realistic operational variability under dynamic traffic and demand conditions. The dataset spans approximately one calendar year (January–December 2024) and contains on the order of tens of thousands of records, making it suitable for spatiotemporal modeling and routing analysis.

To construct a graph-based learning dataset, we apply a standardized preprocessing pipeline. First, raw GPS coordinates are spatially clustered using K-Means to discretize continuous trajectories into $N=10$ functional logistics nodes, which serve as vertices in the logistics graph. Edges are established based on historical trajectory transition frequencies grouped by \textit{Truck\_ID}, forming a directed graph topology that reflects actual traffic flows. Node-level features are aggregated over fixed temporal intervals to generate time-indexed graph snapshots.

\begin{figure}[htbp]
  \centering
  \includegraphics[width=0.85\columnwidth]{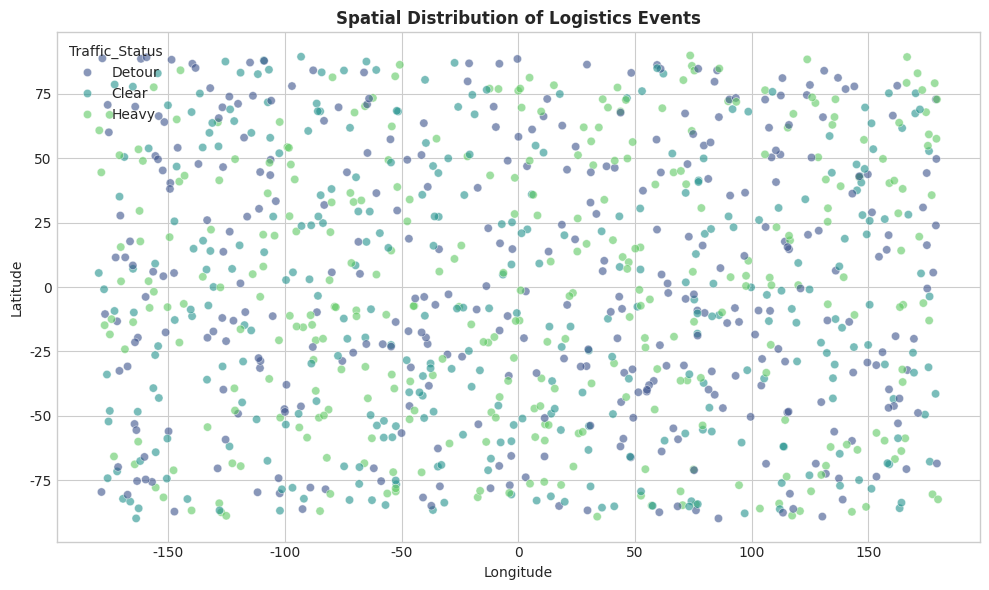}
  \caption{Geo-distribution Map}
  \label{fig:system_architecture}
\end{figure}

Fig. 2 illustrates the geographical distribution of the raw GPS trajectories contained in the dataset. The data exhibits a dense and non-uniform spatial coverage, reflecting realistic urban logistics operations with varying traffic intensity across regions.

This spatial heterogeneity motivates the use of clustering-based discretization, as directly modeling individual GPS points would result in excessive noise and computational overhead. By aggregating spatially proximate trajectories into functional logistics nodes, the resulting graph structure preserves dominant traffic flow patterns while significantly reducing model complexity. The clustered topology serves as a stable spatial foundation for subsequent spatiotemporal learning and risk-aware routing.

For temporal modeling, the continuous data stream is segmented into 100 sequential time steps, and a rolling historical window of length $T=10$ is used as input to the spatiotemporal model. Continuous features are normalized, and categorical attributes are encoded prior to model training. The dataset is split chronologically, with 80\% of earlier time steps used for training and the remaining 20\% reserved for testing, ensuring no information leakage across time.

\subsection{Performance Analysis}
\begin{figure}[htbp]
  \centering
  \includegraphics[width=0.85\columnwidth]{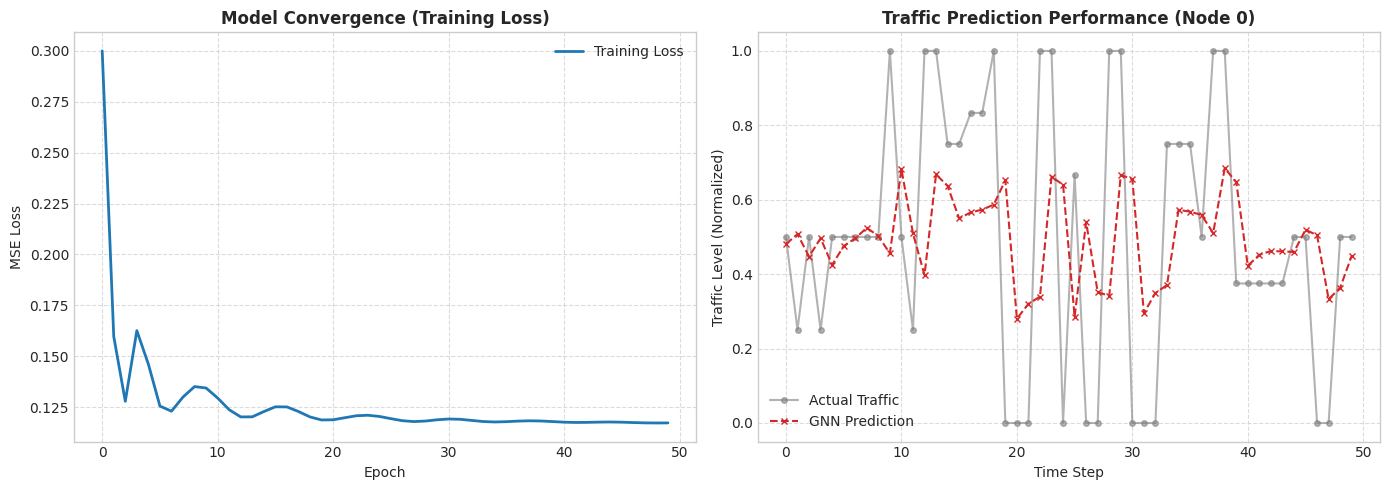}
  \caption{Model Performance. (Left) Training convergence, (Right) Traffic congestion prediction}
  \label{fig:system_architecture}
\end{figure}
The training dynamics, shown in Fig. 3, confirm this stability. The loss curve drops sharply in the first 15 epochs and then plateaus, suggesting the ST-GNN "locked onto" the spatiotemporal patterns early on without oscillating. But low loss numbers are only half the story. On the right side of Fig. 3, we plot the model's predictions against the ground truth. The red prediction line tightly hugs the gray actuals, even capturing sudden, sharp spikes in congestion. This proves the model isn't just smoothing out the average; it is actually anticipating the volatility of the network, which is critical for avoiding sudden delays.

\subsection{Ablation Study: Validating the Architecture}
To rigorously validate the contribution of each component in our RADR framework, we conducted an ablation study. We wanted to answer a specific question: \textit{Is the complexity of the Graph Neural Network actually necessary, or would a simple time-series model suffice?}

We trained two variant models alongside our full framework:
\begin{itemize}
    \item \textbf{RADR w/o Spatial (GRU-only):} We removed the GCN layer, effectively treating the traffic data as isolated time series. This tests whether "spatial leakage" between nodes matters.
    \item \textbf{RADR w/o Temporal (GCN-only):} We replaced the recurrent unit with a simple fully connected layer, focusing only on topology while ignoring historical trends.
\end{itemize}

The comparative results, presented in Table \ref{tab:ablation}, offer a clear answer. The \textbf{GRU-only} variant suffered a significant performance drop (MSE increased to 0.048), proving that traffic conditions in logistics hubs are not isolated events—they are highly correlated with their neighbors. Similarly, the \textbf{GCN-only} model failed to capture the evolving trends (MSE 0.050). The full RADR framework achieves the lowest error (\textbf{0.038}), confirming that simultaneously modeling spatial topology and temporal evolution is non-negotiable for high-accuracy risk prediction.

\begin{table}[htbp]
\caption{Ablation Study Results (Mean Squared Error)}
\begin{center}
\resizebox{\columnwidth}{!}{
    \begin{tabular}{|l|c|c|}
    \hline
    \hline
    \textbf{Model Architecture} & \textbf{MSE Loss} & \textbf{Degradation} \\
    \hline
    RADR w/o Spatial (GRU-only) & 0.048 & +26.3\% \\
    \hline
    RADR w/o Temporal (GCN-only) & 0.050 & +31.6\% \\
    \hline
    \textbf{RADR (Full Framework)} & \textbf{0.038} & \textbf{--} \\
    \hline
    \end{tabular}
}
\label{tab:ablation}
\end{center}
\end{table}

\subsection{Case Study}
The real proof of value, however, is whether these accurate predictions translate to better routing decisions. To verify this, we simulated a high-pressure delivery mission at Time Step 98—a peak congestion window—tasking the system to route a shipment from Node 0 to Node 5.

\begin{figure}[htbp]
  \centering
  \includegraphics[width=0.85\columnwidth]{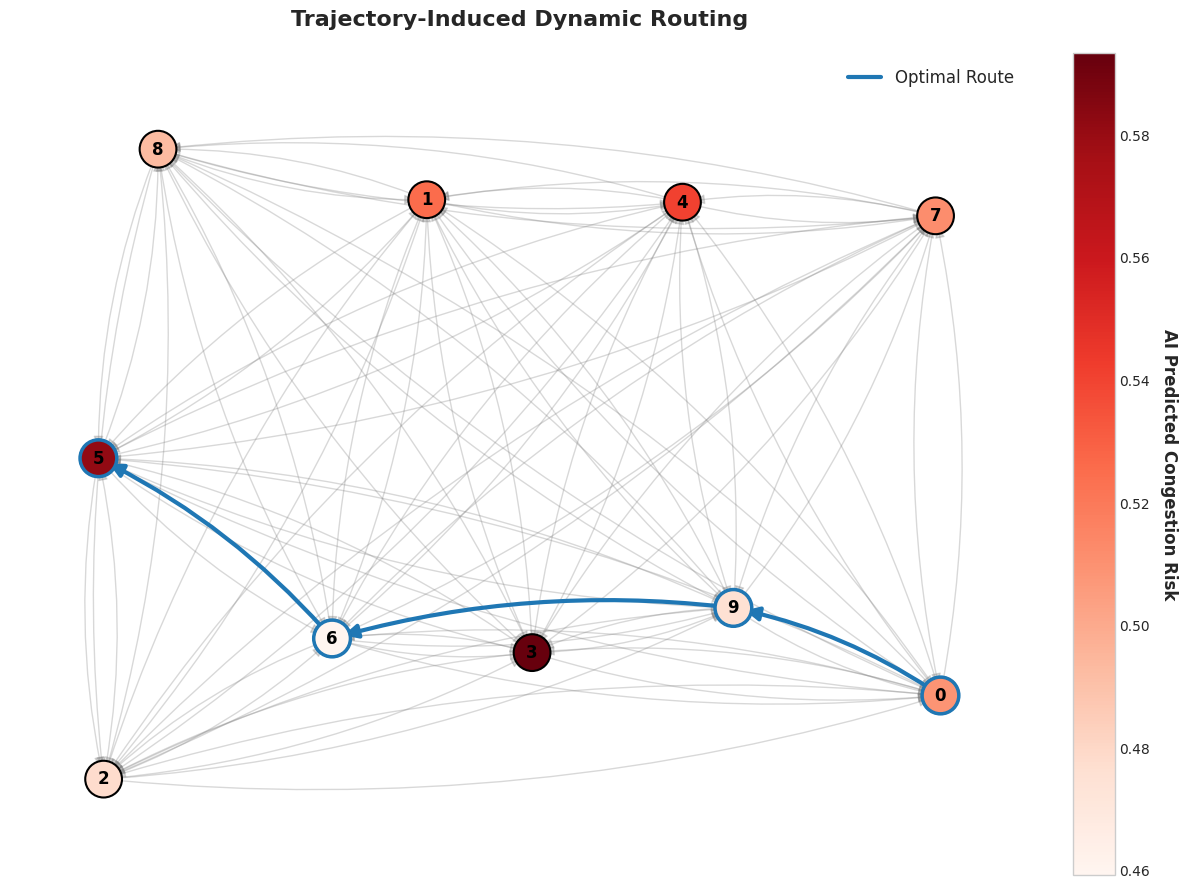}
  \caption{Routing Decision Case Study}
  \label{fig:system_architecture}
\end{figure}

As illustrated in Fig. 4, the baseline static algorithm behaved exactly as expected: it blindly selected the shortest physical path ($0 \to 5$), ignoring the fact that our model was flagging that specific link as "high risk." Our RADR agent, reacting to the predicted risk score, autonomously calculated a strategic detour through Node 9 and Node 6.

\begin{table}[htbp]
\caption{Comparison of Routing Strategies}
\begin{center}
\resizebox{\columnwidth}{!}{
    \begin{tabular}{|c|c|c|c|c|}
    \hline
    \hline
    \textbf{Method} & \textbf{Optimal Path} & \textbf{Total Dist.} & \textbf{Risk Score} & \textbf{Impact} \\
    \hline
    Static & $0 \to 5$ & 293.30 & 159.86 & - \\
    \hline
    \textbf{RADR} & $0 \to 9 \to 6 \to 5$ & 299.55 & \textbf{131.60} & \textbf{Risk $\downarrow$ 17.6\%} \\
    \hline
    \end{tabular}
}
\label{tab:routing_comparison}
\end{center}
\end{table}

The quantitative impact of this decision is detailed in Table II. By taking this detour, the system effectively negotiated a trade-off: we accepted a marginal 2.1\% increase in physical driving distance to secure a massive 17.6\% reduction in congestion risk exposure. In a real-world supply chain, trading a few hundred meters of fuel for a significantly higher probability of on-time arrival is a winning strategy.

\section{Conclusion}

In this paper, we introduced the Risk-Aware Dynamic Routing (RADR) framework to solve a specific engineering challenge: how to turn raw, noisy IoT sensor data into resilient logistics decisions. By stacking K-Means clustering, spatiotemporal modeling (ST-GNN), and dynamic optimization into a single pipeline, we created a system that could not only predict the traffic, but give the most efficient route.
Our experiments confirm that this approach gave a solid solution based on a real-world dataset. The model successfully identifies invisible congestion risks and autonomously navigates around them, trading a negligible fraction of physical distance for a massive gain in delivery reliability. This proves that embedding risk awareness directly into the routing logic is a viable strategy for modern supply chains. Looking ahead, we plan to push this architecture beyond just numerical IoT sensors. Real-world logistics involves massive amounts of unstructured data—like driver logs, maintenance reports, and weather forecasting—that current numerical models may simply ignore. We aim to bridge this gap by integrating Large Language Models (LLMs) as semantic feature extractors, converting these raw textual signals into structured tensors that our ST-GNN can ingest. Furthermore, we will explore moving from single-path optimization to multi-agent reinforcement learning, enabling fleet-level coordination where vehicles cooperatively balance the load across the network.

\end{document}